\documentclass[letterpaper]{article}
\pdfoutput=1  
\usepackage{aaai16}
\usepackage{times}
\usepackage{helvet}
\usepackage{courier}
\usepackage{graphicx,caption,subfigure}
\usepackage{amsmath,amssymb,bm,upgreek,mathrsfs}
\usepackage{color}
\usepackage{algorithmic,algorithm}
\usepackage{enumitem}
\usepackage{nameref}
\DeclareMathOperator*{\argmax}{arg\,max}

\newcommand{\Norm}{\mathcal{N}}

\newcommand{\CRP}{\mathsf{CRP}}

\newcommand{\given}{\, \vert \,}

\newcommand{\R}{\mathbb{R}}

\newcommand{\N}{\mathbb{N}}
\newcommand{\abs}[1]{\left\vert #1 \right\vert}

\newcommand{\bx}{\mathbf{x}}
\newcommand{\bX}{\mathbf{X}}
\newcommand{\by}{\mathbf{y}}
\newcommand{\bY}{\mathbf{Y}}

\newcommand{\bC}{\mathbf{C}}

\newcommand{\bz}{\mathbf{z}}
\newcommand{\br}{\mathbf{r}}

\newcommand{\bT}{\mathbf{T}}

\newcommand{\cA}{\mathcal{A}}

\newcommand{\cZ}{\mathcal{Z}}
\newcommand{\cR}{\mathcal{R}}

\title{Bayesian Inference of Recursive Sequences of Group Activities from Tracks}
\author{Ernesto Brau\textsuperscript{1}, Colin Dawson\textsuperscript{2}, Alfredo Carrillo\textsuperscript{3}, David Sidi\textsuperscript{3}, Clayton T.~Morrison\textsuperscript{3}\vspace{2mm}\\
\textsuperscript{1}Computer Science Department, Boston College;~~{\footnotesize \tt brauavil@bc.edu} \\
\textsuperscript{2}Department of Mathematics, Oberlin College;~~{\footnotesize \tt cdawson@oberlin.edu} \\
\textsuperscript{3}School of Information, University of Arizona;~~{\footnotesize \tt \{isaac85,dsidi,claytonm\}@email.arizona.edu} }

\begin{document}
\maketitle

\begin{abstract}
\begin{quote}
We present a probabilistic generative model for inferring a description of
coordinated, recursively structured group activities at multiple levels of
temporal granularity based on observations of individuals' trajectories. The
model accommodates: (1) hierarchically structured groups, (2) activities that
are temporally and compositionally recursive, (3) component roles assigning
different subactivity dynamics to subgroups of participants, and (4) a
nonparametric Gaussian Process model of trajectories. We present an MCMC
sampling framework for performing joint inference over recursive activity
descriptions and assignment of trajectories to groups, integrating out
continuous parameters. We demonstrate the model's expressive power in several
simulated and complex real-world scenarios from the VIRAT and UCLA
Aerial Event video data sets.
\end{quote}
\end{abstract}


\section{Introduction}
\label{sec:intro}

Human activity recognition comprises a range of open challenges and is a very
active research area \cite{aggarwal2011,vishwakarma2013,sukthankar2014},
spanning topics from visual recognition of individual behavior
\cite{poppe2010}, pairwise interactions among individuals participating in different
roles in a joint activity \cite{barbu2012,9_kwak2013}, coordinated sequences
of actions as expressions of planned activity \cite{geib2009}, and multiple
groups of individuals interacting across broad time scales.  In this paper, we
address the last of these, presenting a framework for automatically
constructing an interpretation of high-level human activity structure as
observed in surveillance video, across multiple, interleaved instances of
activities.  We assume that lower-level visual processing provides high quality
tracks of individuals moving through the scene.  Our goal is to construct
accurate descriptions of the events in the video at different levels of
granularity, based on the tracks alone.  We develop a probabilistic generative
model that combines multiple features that to our knowledge have not been
previously incorporated into a single framework for joint inference.  To wit:
(1) Activities have {\em composite structure} with {\em roles} representing
    semantically distinct aspects of the overall activity structure.
(2) Activities are described {\em hierarchically} and {\em recursively},
    entailing {\em multiple levels of
      granularity} both in time and membership.  
(3) Arbitrarily sized groups of actors participate in activities and fulfill
    roles.  
(4) Hierarchical descriptions and temporally changing groupings consist of
    the {\em best joint explanation} of the full set of individual
    trajectories, as found via posterior probabilistic inference.



The rest of the paper is organized as follows.  In the next section, we review
prior research, with a focus on work modeling group membership, hierarchically
structured activities, and the identification of roles.  In
Section~\ref{sec:model} we present our probabilistic generative
model.  In
Section~\ref{sec:inference} we present an MCMC sampling framework for
performing joint inference using the model.  In Section~\ref{sec:experiments} we
evaluate the model on synthetic and real-world data from the VIRAT 
\cite{42_oh2011large} and UCLA Aerial Event \cite{shu2015_CVPR} video data sets, 
demonstrating the model's expressive power and effectiveness.
We conclude with a discussion of future work.


\section{Related Work}

\begin{figure*}[t!]
\centering
\subfigure[Example activity tree]{
    \includegraphics[width=0.51\linewidth, height=6cm]{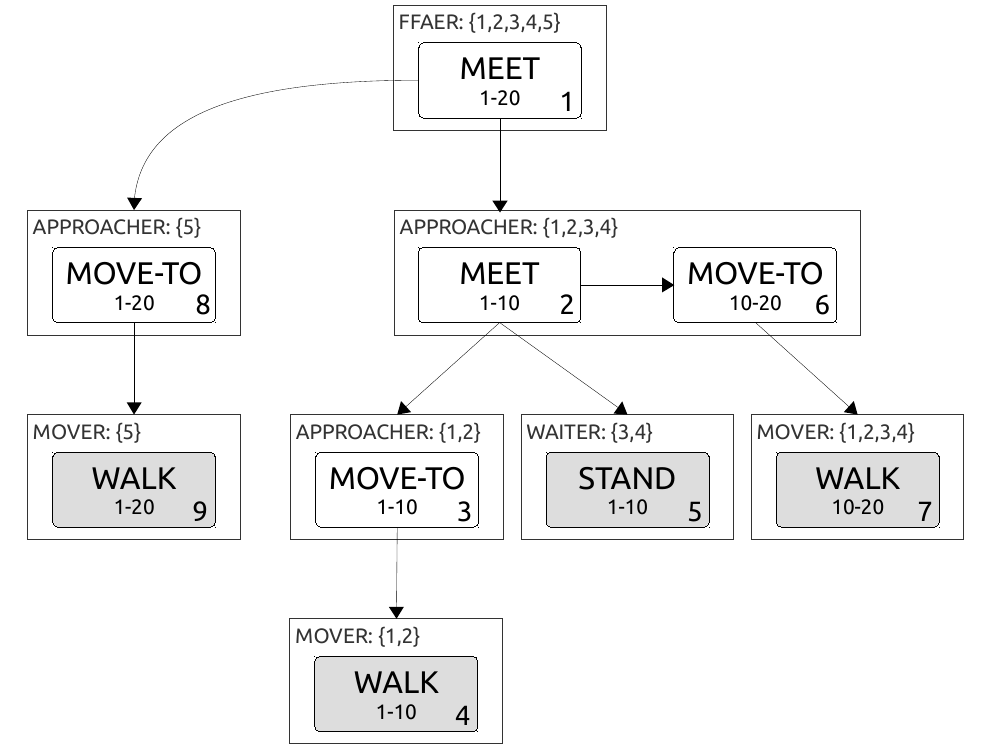}
    \label{fig:ex-tree}
}
\subfigure[$t = 5$]{
    \includegraphics[width=0.22\linewidth, height=6cm]{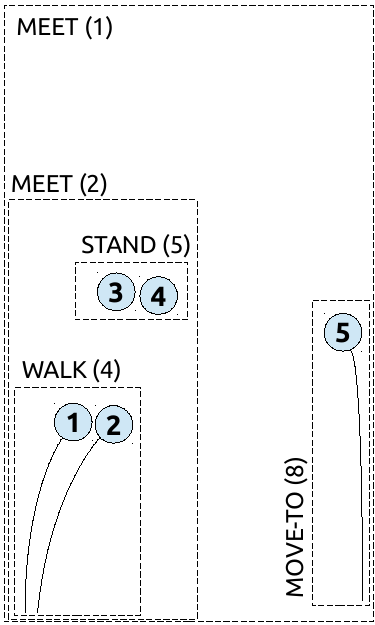}
    \label{fig:ex-trajs-1}
}
\subfigure[$t = 20$]{
    \includegraphics[width=0.22\linewidth, height=6cm]{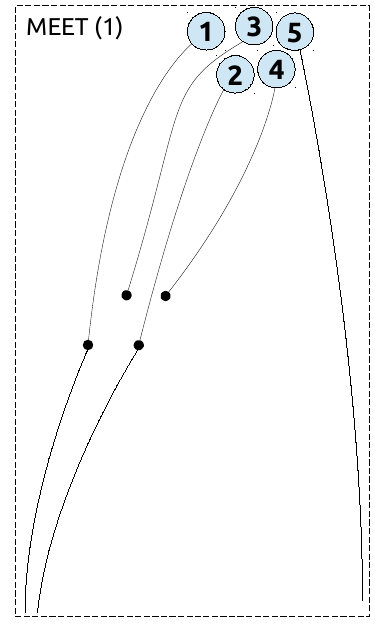}
    \label{fig:ex-trajs-3}
}
\caption{This example depicts five synthetic trajectories and the corresponding
    activity tree representation. 
    Each node in the activity tree \subref{fig:ex-tree}~
    represents an activity.  The central label
    indicates the activity name, with the time interval for the activity
    underneath. Each node is also numbered with a unique ID, which is displayed
    in the bottom-right corner of each node rectangle. Shaded nodes indicate
    physical activities.
    The outer rectangles represent \emph{activity sequences},
    whose roles and set of participants are displayed in the top left
    corner of their rectangles. For example, node $2$ is activity $C_2$ with
    label $a_2 = \text{{\sc meet}}$, start time $s_2 = 1$, and end time $e_2 =
    10$.
    Node 2 is part of an activity sequence
    realizing the role of {\sf APPROACHER} by individuals $\cZ_2 = \{1,2,3,4\}$.
    In \subref{fig:ex-trajs-1}~ and \subref{fig:ex-trajs-3}~ we show two
    frames of the corresponding scene.
    Each circle represents an individual ($1$--$5$), and the curves
    are their paths on the ground plane. The dotted rectangles represent
    activities performed by the individuals inside them, labeled with
    their activity name and activity tree node number.
    For example, we can see in \subref{fig:ex-trajs-1}~, at time $t = 5$,
    individuals $1$ and $2$ are walking together (node $4$)
    to meet (node $2$) with individuals $3$ and $4$, who are standing (node $5$).
    All this happens while all individuals are participating in another
    global meeting together (node $1$).  By frame $t = 20$, depicted in
    \subref{fig:ex-trajs-3}~, 
    all participants are meeting.
    (Note: to simplify the presentation here, we omit the root {\sc FFA}
    activity from the example activity tree in \subref{fig:ex-tree}~.)}
\label{fig:example}
\end{figure*}

A number of researchers have proposed models that distinguish the different
roles that individuals play in a coordinated activity
\cite{2_ryoo2011,barbu2012,lan2012,9_kwak2013}.
These models capture the semantics of activities
with component structure.  It can be difficult to scale role identification in
scenes with an arbitrary number of individuals, in particular while properly
handling identification of non-participants \cite{9_kwak2013}.  A consequence
of our joint inference over role assignments and groups is that our model
naturally distinguishes and separates participants of different activities
playing different roles.

Considerable work has been devoted to developing more expressive models in
which activities are decomposed into hierarchical levels of description, across
different spatial and temporal granularities.  Some prior models account for a
specific number of hierarchical levels of description, with up to 3 levels
being a popular choice
\cite{1_choi2012,4_lan2010beyond,17_chang2011probabilistic,6_cheng2014recognizing}.
Other models permit a potentially greater number of levels of activity
description, but the activity hierarchy is fixed prior to inference
\cite{9_kwak2013,14_garate2014group,23_zaidenberg2012generic}.  In only a few
cases, including our model, are the levels of activity description assessed
during inference \cite{3_lin2010group,2_ryoo2011}.

A third branch of work has been devoted to modeling activities not just among
individuals, but involving groups of actors.  In some models, activities
include groups, but interactions are still considered between individuals
within groups
\cite{1_choi2012,4_lan2010beyond,28_zhang2013beyond,33_odashima2012collective,36_zhu2011generative}.
Other models allow for activities to be performed between groups themselves
\cite{17_chang2011probabilistic,18_chang2010group}.  
Still others, including our model, take group activity modeling a step further,
allowing for arbitrary numbers of participants in groups, provided they satisfy
group membership criteria while performing the activity
\cite{3_lin2010group,9_kwak2013,shu2015_CVPR,24_zhang2012parsing}.


Two recent papers, one by \citeauthor{2_ryoo2011}
\shortcite{2_ryoo2011} and one by \citeauthor{shu2015_CVPR}
\shortcite{shu2015_CVPR}, come closest to accommodating the
combination of features in our model. Both have developed methods for
simultaneous inference of group structure with roles in hierarchically
structured activities. One key difference between the two is that the
former, like our model, can flexibly accommodate multiple levels of
activity description in the course of inference, while the latter is
restricted to two fixed levels of description during inference
(one for group activities, and the other for their member
roles). 
A crucial difference with our approach is that we consider the possibility of
detecting multiple occurrences of the same type of activity description in a
scene, and these descriptions are integrated into a single probabilistic joint
model in order to influence inference of the overall activity structure.  
While activities described in \citeauthor{2_ryoo2011} \shortcite{2_ryoo2011}
have hierarchical component structure with roles, their model does not
accommodate more than one top level activity (i.e. assault, meet, robbery,
etc.) occurring during the course of the video, nor changes in group membership
by individuals over such high-level activities.


\section{Model}
\label{sec:model}

We present a probabilistic generative model describing how
coordinated activities by groups of individuals 
give rise to the observed physical trajectories of the actors involved.  In
Section \ref{sec:terms} we
introduce some terminology.  Then, in Section \ref{sec:representation} 
we give precise definitions of the
representations employed by our model for activities, groups, activity
sequences, and spatial trajectories.  Next, in Section
\ref{sec:topology}, we define the factors in the joint probability
distribution that comprise the generative model.  Finally, in
Section \ref{sec:specific-activities} we define the specific
activities that are used to model the scenes that we use for
evaluation in Section \ref{sec:experiments}.

\subsection{Terminology}
\label{sec:terms}

{\em Activities} are optionally composed of other activities.  For example,
the canonical {\sc meet} activity in our model consists of multiple
participating groups, each of which must {\sc move-to} the location 
of the meeting, unless they are already there, and then possibly wait ({\sc
stand})
until one or more other groups arrive.  Engaging in a {\sc move-to} activity
requires the group to carry out a sequence of {\sc walk} and {\sc run}
activities. The semantics of a {\sc walk} or {\sc run} is, in turn,
characterized by
the group's physical trajectory in space.  

A scene is
described as an {\em activity tree}, which has a recursive structure,
much as a syntactic tree describes the recursive phrase structure of a
natural language sentence, but with an added temporal component.
An example of such a tree is depicted in Figure~\ref{fig:ex-tree}.
We refer to the more abstract, ``nonterminal'' activities that are defined
in terms of 
other activities as {\em intentional}, and the ``leaf'' activities that are
defined in terms of observable movement as {\em physical}.

\emph{Participants} in an intentional activity are divided into
\emph{subgroups}, each of which plays a particular \emph{role} with respect
to the parent activity.  Carrying out a particular role entails
engaging in a sequence of subactivities, each of which may be physical
or intentional.  In intentional activities, the group may be decomposed
further into smaller subgroups, each with their own role in the
subactivity, and so on.  The \emph{sequence} of subactivities
performed by participants in a role is constrained (either
deterministically or stochastically) according to the dynamics associated with
the role.
In cases where a role in a parent activity
may be realized as a sequence that includes the same activity type as
the parent, the structure is recursive, and allows for arbitrary
nesting of activities at different time scales. For example, in Figure
\ref{fig:example},
a group of individuals participating in a {\sc meet} activity
performs a sub-{\sc meet} of their own.

Physical activities are associated with \emph{group trajectories},
which are coupled either via shared membership or
when an intentional activity requires their coordination.  
An individual, $j$, has an observed trajectory that is generated as a sequence
of connected sub-trajectories, one for each physical activity in which $j$
participates, and which is constrained to be near the respective group
trajectory.
An example of an observed set of individual trajectories is shown in Figure
\ref{fig:ex-trajs-3}.  

An important feature of our model is
that group trajectories are not explicitly represented during
inference. The assertion that there is some group trajectory induces
correlations among the individual trajectories of the members, but we
average over all possible group trajectories (marginalizing them out) 
when computing the posterior probability of a description.  This allows
activity descriptions with different numbers of groups to be compared based on
the posterior probabilities of their activity trees alone, without
needing to deal with probability densities with different numbers of dimensions.


\subsection{Representation}
\label{sec:representation}


\subsubsection{Activities}

Formally, an activity is a tuple $C = (a, s, e)$, where
$a$ is an activity label (e.g., {\sc walk}), and $s, e \in \mathbb{R}^{+}$
are the start and end time of the activity, respectively.
The simplest activities are \emph{physical} activities, e.g.,
walking, running and standing, which directly constrain the motion of a group
of individuals
over an interval of time. For example,
a {\sc run} activity is expected to yield trajectories with speeds
corresponding to typical human running.
We denote the set of physical activity labels by
$\cA_{\mathrm{phys}}$. Similarly, \emph{intentional} activities
include {\sc meet} and {\sc move-to}. We denote the set of intentional
activity labels by $\cA_{\mathrm{int}}$.  The complete set of activities is
denoted by
$\cA = \cA_{\mathrm{phys}} \cup \cA_{\mathrm{int}}$.

\subsubsection{Groups}

The set of participants in activity $C$ is denoted by $\cZ_C
\subset \N$.  We let $J_C = \abs{\cZ_C}$, the size of the group.
This set is partitioned into subgroups, $\{\cZ_1, \dots, \cZ_{K_C}\}$,
where the number $K_C$ of subgroups is bounded only by the number of
individuals, $J_C$, in the group.  When specifying the probability
model, it is convenient to work instead with the
indicator variables, $\bz_C = (z_{C1}, \dots, z_{C J_C})$, where $z_{Cj}$
indicates which
of the $K_C$ subgroups participant $j$ is affiliated with.  Note that
$\bz_C$ contains exactly the same information as the partition.
In the following description, we omit the subscript $C$ for readability.

\subsubsection{Activity Sequences}

Each subgroup within activity $C$ performs a {\em sequence} of
subactivities.  For example, a subgroup in a {\sc meet} might perform
a sequence of two subactivities: first, {\sc move-to} a designated
meeting location, and then {\sc stand} in that location while meeting
with other subgroups, who may have also approached that location.
Alternatively, one of the subgroups could be involved in a side meeting,
with further subgroups that approach and merge with each other before their
union merges with another subgroup of the top-level {\sc meet}.
Figure~\ref{fig:example} provides such an example. 

The sequences of subactivities performed by subgroups are governed by
{\em roles}.  Each subgroup $k$ is assigned a role from a set $\cR_a$
defined by
the parent activity $a$.  Roles govern the dynamics of the sequence of 
subactivities that each group carries out, via a set of parameters
associated with that role.  The parameters for role $r$ specify what the
allowable activities are in the activity sequence of a group carrying
out that role, as well as hard or soft constraints involving what
order the activities occur in.
The subgroup of the example {\sc meet} activity above would be assigned the
role of {\sf APPROACHER}, which prescribes a {\sc move-to} followed by a
{\sc stand}.  In this case the constraint on the order of
subactivities is deterministic, with the only degrees of freedom
being the times at which transitions take place.  These constraints
can be represented by a Markov chain with a degenerate initial distribution and a
transition matrix with only one non-zero off-diagonal entry per
column.  More general initial and transition distributions will give
rise to softer constraints on activity sequences.

\subsubsection{Trajectories}

Ultimately, each physical activity, realized over the interval
$[s,e]$, is associated with a {\em group
trajectory}, denoted by $\bx$, which is a 2 by $e-s+1$ array specifying
a 2-dimensional position on the ground plane for each time index
between $s$ and $e$ inclusive.
The group trajectory represents the central tendency
of the members' individual trajectory segments during the interval $[s,e]$.
Since individual $j$'s path depends on
the sequence of activities it participates in, each
individual trajectory $\by_j$ consists of segments, $\by_j^{(0)}, \dots,
\by_j^{(I_j)}$, consecutive pairs of which must be connected
at {\em transition points}, $y_{*j}^{(1)},
\dots, y_{*j}^{(I_j)}$, where $y_{*j}^{(i)}$ denotes the start of segment $i$
and the end of segment $i-1$.

\subsection{Generative Model}
\label{sec:topology}

We now describe the generative process for
activities.  The high-level process has three steps: (1) recursive 
expansion of intentional activities, (2) generation of group
trajectories for the set of physical activities, and (3) generation of
individual trajectories conditioned on the group assignments and group
trajectories.

\subsubsection{Overview}

In the first step of activity generation, each intentional activity gives rise to one or more
child activity
sequences: one for each subgroup of participants involved in the
parent activity.  Each child sequence is assigned a role, based on
the parent activity type.  Subgroups and role assignments occur
jointly.  The choice of role governs the sequence of 
activities that the subgroup engages in, by specifying a Markov transition function. 
Each segment of an activity sequence may be a physical
activity or another intentional activity.  
Each intentional activity in the sequence is recursively expanded until only
physical activities are generated.

As a working example for this stage of the process, 
we consider the {\sc meet} activity at node 1,
at the root of the tree in Figure~\ref{fig:ex-tree}.  Node 1 has two child
sequences corresponding to the two subgroups involved in the meeting, 
both carrying out the {\sf APPROACHER} role.  One of those child sequences
consists of just a
single {\sc move-to} activity, while the other consists of two
activities: a {\sc meet}, followed by a {\sc move-to}.  
In general, a special top-level root activity, a ``free-for-all'' ({\sc FFA}),
comprises all actors and has a duration over the entire video.  All other
activities are children of the root {\sc FFA}.  (To simplify the example tree
in Figure~\ref{fig:ex-tree}, we removed the parent {\sc FFA}.)
The details of this tree expansion are given in
``\nameref{sec:activity-tree-generation}'',
below.

We make a conditional independence assumption by supposing that the
contents 
of a parent activity fully specify the distribution of possible child 
sequences, and that child sequences are conditionally independent 
of each other given their parent.  Since child sequences can contain 
other intentional activities, activity generation is a recursive
process, which bottoms out when no intentional activities are
generated.

In the second step of the generative process, group trajectories are generated for each physical activity.
This process must satisfy two constraints:
(a) physical activity trajectories that share members and that border in time
must be spatially connected;
and
(b) groups that need to physically interact as co-participants in an
intentional activity, such as a {\sc meet}, must have trajectories that
intersect at the appropriate points in time.  
Due to these constraints
it is not feasible to generate group trajectories conditionally
independently given the activity tree.  Instead they are generated
jointly according to a global Gaussian Process with a covariance
kernel that depends on the activity tree in such a way as to enforce
the key constraints.  The details of this process are given in
``\nameref{sec:group-trajectory-generation}'',
below.

For the example tree in Figure~\ref{fig:ex-tree}, four group trajectory
segments are needed, one
for each physical activity leaf node.  Since {\sc walk} activity 9 is part
of a meeting in which its participants must meet with the participants
in {\sc walk} activity 7, the group trajectories for 7 and 9 must end in the
same location.  Similarly, {\sc walk} 4 must end where {\sc stand} 5
is located. 

In the final step of the generative process, the individual trajectories are realized, conditioned on the
set of group trajectories.  Here, conditional independence is possible,
with each individual's trajectory depending only on the sequence of
group trajectories for physical activities in which that individual is
a participant.  This process is detailed in Section
``\nameref{sec:individual-trajectory-generation}''.


\subsubsection{Generating the Activity Tree}
\label{sec:activity-tree-generation}

Let $C_m = (a_m, s_m, e_m)$ be a parent intentional activity,
where $m$ indexes the set of activities. Its
participant set $\cZ_m$ (by relabeling, we assume
that $\cZ_m = \{1, \dots, J_m\}$) is divided into subgroups,
where the $j$th participant of $C_m$ is assigned to group $z_{mj}$,
and the distinct realized groups are numbered $1$ through $K_m$.
We let $\bz_m = (z_{m1}, \dots, z_{m J_m})$, which defines a partition
of $\cZ_m$ into $K_m$ subgroups, $\{\cZ_{m1}, \dots, \cZ_{m K_m}\}$,
with $\cZ_{mk} = \{j \in \cZ_m \given z_{mj} = k\}$.
Subgroup $k$ is assigned role $r_{mk} \in \cR$, and we define
$\br_m = (r_{m1}, \dots, r_{m K_m})$.

\begin{figure}
    \centering
    \includegraphics[width=\linewidth]{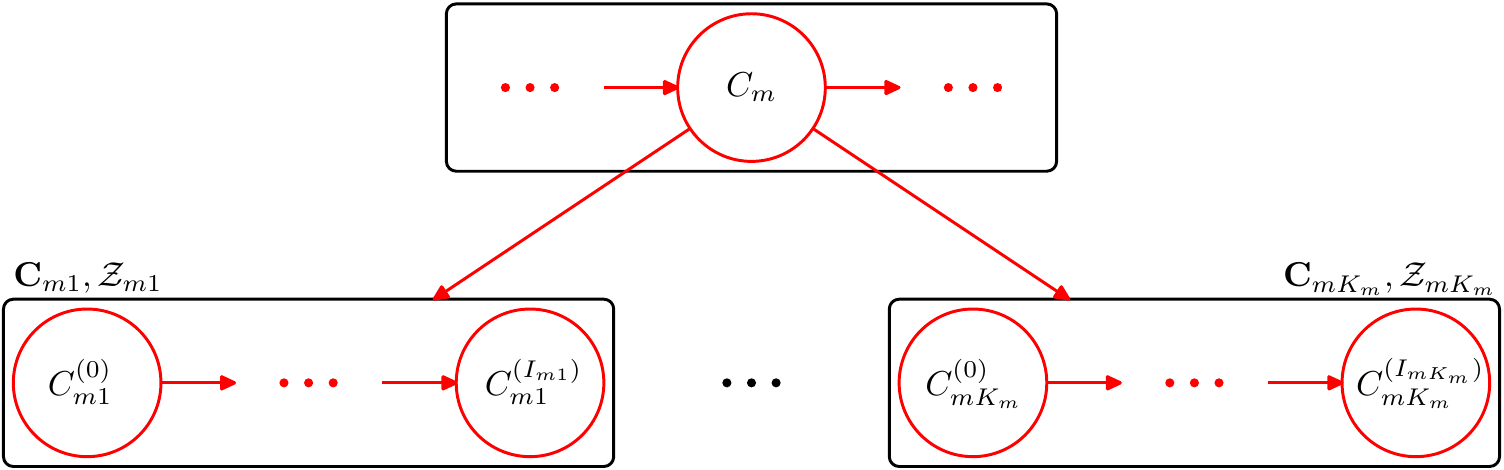}
    \caption{Graphical model for an intentional activity. The set $\cZ_m$ of
        participants of activity $C_m$ is partitioned into groups
        $\cZ_{m1}, \dots \cZ_{m K_m}$ via the indicator vector
        $\bz_m$.  Each group is assigned a role, $r_{mk}$, and performs an
        activity
        sequence $\bC_{mk} = (C^{(0)}_{mk}, \dots, C^{(I_{mk})}_{mk})$,
        $k = 1, \dots, K_m$.}
    \label{fig:activity}
\end{figure}

The $k$th subgroup, which has participants $\cZ_{mk}$ and role $r_{mk}$,
produces an activity sequence according to the stochastic process
associated with $r_{mk}$, which has parameters
$\boldsymbol{\pi}_{r_{mk}}$, a Markov transition function, and $\bT_{r_{mk}}$, an initial activity distribution.
Denote the resulting sequence by
$\bC_{mk} = (C_{mk}^{(0)}, \dots, C_{mk}^{(I_{mk})})$, where $I_{mk}$ is
the number of jumps generated by the process, and the $C^{(i)}_{mk}$ are
activity tuples, $C_{mk}^{(i)} = (a_{mk}^{(i)}, s_{mk}^{(i)}, e_{mk}^{(i)})$,
where $s_{mk}^{(i)} = e_{mk}^{(i-1)}$. Figure \ref{fig:activity} illustrates
the graphical model of this production.

To summarize, the grouping, role assignment, and child activity
sequences within activity $C_m$ are generated according to
$p(\bz_m, \br_m, \bC_m \given C_m, \cZ_m)$, which factors as
\begin{align}
\label{eq:sequence}
p(\bz_m \given a_m, \cZ_m) p(\br_m \given a_m)
    \prod_{k=1}^{K_m} p(\bC_{mk} \given
        s_m, e_m),
\end{align}
where we define $\bC_m = (\bC_{m1}, \dots, \bC_{m K_m})$. We assume that
roles are assigned independently, so that
$p(\br_m \given a_m) = \prod_{k=1}^{K_m} p(r_{mk} \given a_m)$.
Additionally, we model the partition of $\cZ_m$ into subgroups using a
Chinese Restaurant Process ($\CRP$), whose concentration parameter
$\alpha_{a_m}$ depends on the activity label.
That is, we let $p(\bz_m \given a_m, \cZ_m)$ be the $\CRP$ mass function
with parameter $\alpha_{a_m}$.
Each segment $C_{mk}^{(i)}$ of $\bC_{mk}$ that is an intentional
activity is expanded and its members are subdivided recursively according
to \eqref{eq:sequence}, replacing $C_m$ with $C_{mk}^{(i)}$ and
$\cZ_m$ with $\cZ_{mk}$. Once all expansions contain
exclusively physical activities, the recursion has bottomed out. The
resulting tree consists of all intentional activities $C_1, \dots,
C_M$, all physical activities, $C_{M+1}, \dots, C_{M+N}$, and, for
each intentional activity $m = 1, \dots, M$, its
membership partition $\bz_m$, role assignments $\br_m$,
and subgroup activity sequences $\bC_m$.
We denote this complete tree by $\Lambda$, whose prior distribution is
\begin{equation}
\label{eq:tree-prior}
p(\Lambda) = \prod_{m=1}^{M} p(\bz_m, \br_m, \bC_m \given C_m, \cZ_m).
\end{equation}

\subsubsection{Generating Group Trajectories}
\label{sec:group-trajectory-generation}

The leaves of the activity tree are all physical activities, each of which is
associated with a group trajectory. In general, the endpoints of different
group trajectories are \emph{not} independent (given $\Lambda$), since
they may be constrained to start or end at the same location.
Consequently, we define a joint distribution on all of the group trajectory
endpoints,
and, conditioned on their endpoints, we treat their interiors as
independent.

We model the interiors as realizations of Gaussian processes \cite{gpml2006}
with the squared-exponential kernel function.  This results in trajectories
that are generally smooth, but flexible enough to allow for different kinds
of motion. We use different scale parameters $\sigma_{a_m}$ depending on the
activity $a_m$, which determines the rate of change of the trajectory.


We specify dependencies among the set of trajectory endpoints by 
first defining an undirected weighted graph $\mathcal{G}$ over the endpoints.
We use this graph to construct a constraint matrix over transition
points by interpreting the sum of the weights on the shortest path
between two nodes as distances.  We then apply a
positive semidefinite isotropic covariance kernel point-wise to the
distance matrix to transform the distances into covariances.

\begin{figure}[th!]
    \centering
    \includegraphics[width=0.96\linewidth]{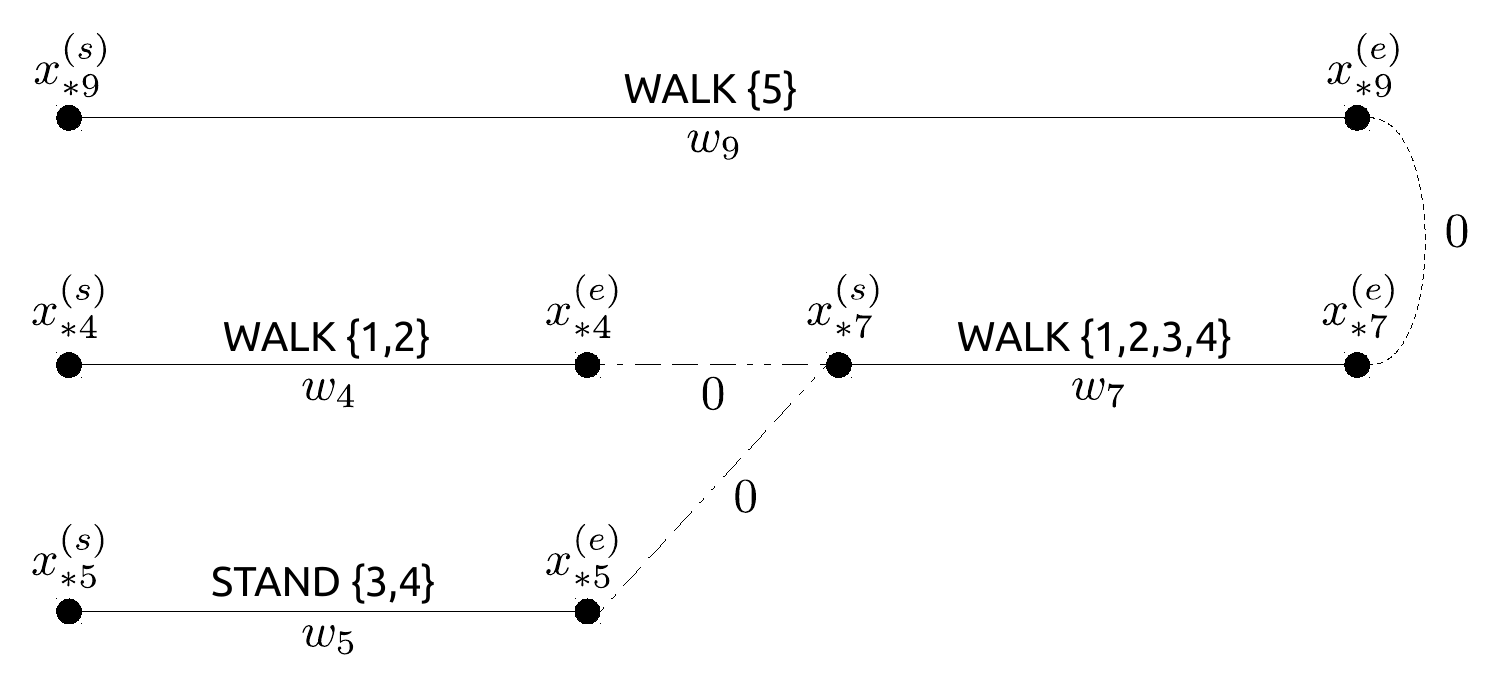}
    \caption{
        Constraint graph for the example activity tree in Figure \ref{fig:example}. The nodes
        in the graph represent endpoints for group trajectories, labeled
        according to the numbering in the tree; e.g., $x_{*9}^{(s)}$
        is the starting endpoint for node $9$. Solid lines represent
        temporal edges, dashed lines are transitional edges, and dotted
        lines are compositional edges (see Section
        \ref{sec:group-trajectory-generation}, ``Generating Group Trajectories''). For clarity, temporal
        edges are labeled with their corresponding physical activity name.
        According to this graph, the distance between $x_{*5}^{(s)}$ and $x_{*9}^{(s)}$
        is $w_5 + w_7 + w_9$, meaning that activities $5$ and $9$ must
        start in locations separated by a {\sc stand} and two {\sc walk}s.}
    \label{fig:ex-graph}
\end{figure}

Let $\bx_m \in \R^{2(e_m-s_m+1)}$ represent the sequence of
ground-plane positions that make up the group trajectory for activity $m$.
Abusing notation slightly, we will write $x_{*m}^{(s)}$ and $x_{*m}^{(e)}$
for both the endpoints of a group trajectory and the
corresponding node in $\mathcal{G}$.  We introduce three kinds of
edges on $\mathcal{G}$: temporal, transitional, and
compositional.  Two nodes are connected by a {\em temporal} edge when they
belong to the same physical activity.  The start of an activity,
$x_{*m}^{(s)}$, is connected to
the end of another activity, $x_{*m'}^{(e)}$, by a {\em transitional} edge
when they
correspond to the same moment in time and the corresponding activities share at
least one participant. 
Finally, two endpoints are connected by a {\em compositional} edge if they
correspond to the same moment in time and have a common ancestor that
specifies they must coincide, e.g., in a {\sc meet} activity, all participants
must \emph{end} in the same location.
All transitional and compositional edges have weight (or ``distance'')
zero, corresponding to the constraint that the connected edges must correspond
to the same
trajectory position.  The weight assigned to temporal edges
is a function of the time elapsed during the
intervening physical activity and the $\sigma$ associated with
the activity label (e.g., ``slower-moving'' activities having lower
weights, corresponding to a stronger dependence between the positions
of their endpoints). 
Figure \ref{fig:ex-graph} shows an example $\mathcal{G}$.

\begin{figure}[h]
    \centering
    \includegraphics[width=0.60\linewidth]{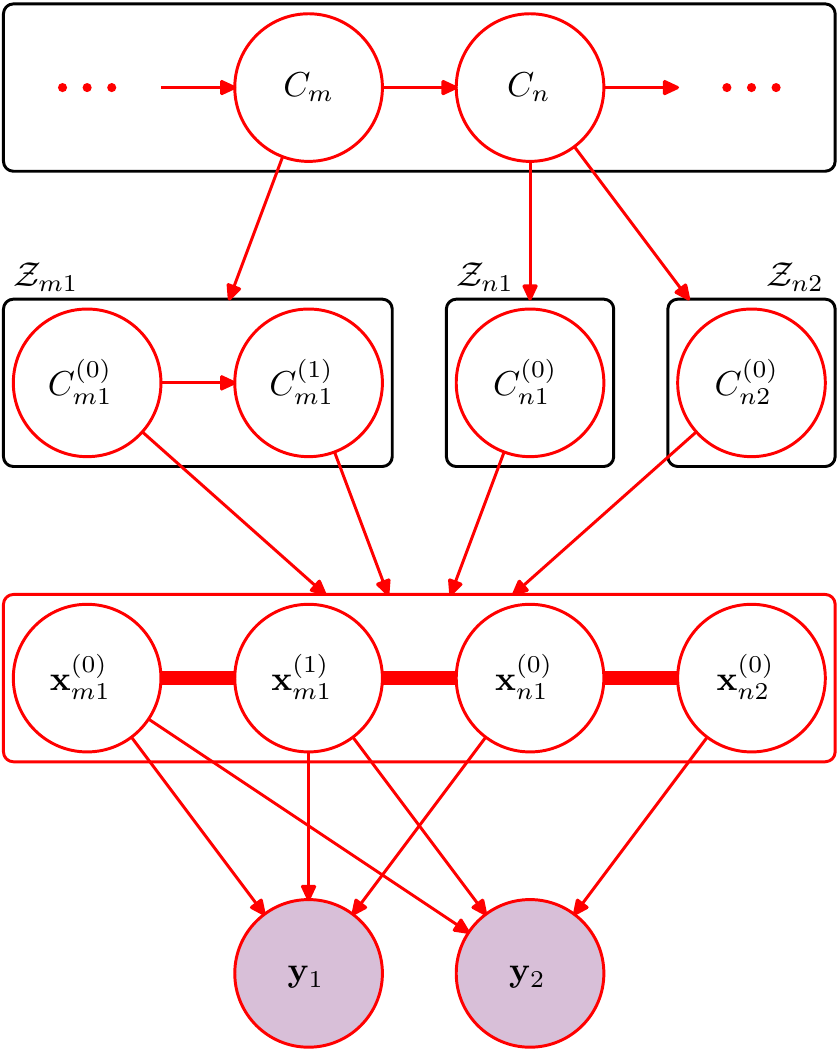}
    \caption{Graphical model for the generation of trajectories.
        This example shows a sequence of activities with two activities
        $C_m$ and $C_n$. $C_m$ has a single child with individuals
        $Z_{m1}$ 
        performing a sequence of two physical activities
        $C_{m1}^{(0)}$ and $C_{m1}^{(1)}$, while $C_n$ is divided
        into two groups, each of which performs a single activity,
        culminating in physical activities
        $C_{n1}^{(0)}$ and $C_{n2}^{(0)}$. Each corresponding group trajectory
        $\bx_{m1}^{(0)}$, $\bx_{m1}^{(1)}$, $\bx_{n1}^{(0)}$, and
        $\bx_{n2}^{(0)}$ potentially depends on \emph{all} physical
        activities, as well as being fully connected with each other
        (represented in the graph by thick red edges). Finally,
        individual trajectories $\by_1$ and $\by_2$ only depend on
        group trajectories whose physical activities contained their
        individuals.}
    \label{fig:act-traj}
\end{figure}

Having defined $\mathcal{G}$ we can compute a distance matrix $\mathbf{D}$ for
the set of physical activity endpoints, where the entry
$d_{ii'}$ is the sum of the weights along the shortest path in
$\mathcal{G}$ from node $i$ to node $i'$. If no path exists between two
nodes, the distance is set to $\infty$.
We then transform $\mathbf{D}$ into a
covariance matrix $\bm{\Phi}$ by applying the covariance function
$\phi_{ii'} = \kappa(d_{ii'}) = \lambda e^{-d_{ii'}^2}$. 
The locations of the set of group trajectory endpoints
$\bX_{*} = (x_{*M+1}^{(s)}, x_{*M+1}^{(e)},
\dots, x_{*M+N}^{(s)}, x_{*M+N}^{(e)})$ is distributed as
$\bX_{*} \given \Lambda \sim \Norm(\mathbf{0}, \bm{\Phi})$.
Conditioned on the endpoints, the interiors
are mutually independent, so that
\begin{equation}
\label{eq:interior-prior}
p(\bX_{-*} \given \bX_{*}) =
\prod_{m=M+1}^{M+N} p(\bx_{-*m} \given x_{*m}^{(s)}, x_{*m}^{(e)}),
\end{equation}
where $\bx_{-*m}$ is the vector of interior points of trajectory $\bx_m$
and $\bX_{-*} = (\bx_{-*M+1}, \dots, \bx_{-*M+N})$.
Each $\bx_{-*m}$ is generated according to the Gaussian process for activity $m$; i.e.,
$\bx_{-*m} \given x_{*m}^{(s)}, x_{*m}^{(e)}$ is normally distributed.
Finally, the distribution over the physical trajectories
$\bX = (\bx_{M+1}, \dots, \bx_{M+N})$ factorizes as
\begin{equation}
\label{eq:physical-traj-prior}
p(\bX \given \Lambda) = p(\bX_{-*} \given \bX_{*}) p(\bX_{*} \given \Lambda).
\end{equation}
Using the fact that factors in \eqref{eq:physical-traj-prior} are normally
distributed, we can easily see that $\bX$ also has a normal distribution.

\subsubsection{Generating Individual Trajectories}
\label{sec:individual-trajectory-generation}

As described above, individual $j$
participates in physical activity sequence $\bC_{(j)}$, which has the
sequence of group trajectories
$\bX_{j} = (\bx_{j}^{(1)}, \dots, \bx_{j}^{(I_j)})$.
The individual trajectory $\by_j$ consists of segments,
$\by_j^{(1)}, \dots, \by_j^{(I_j)}$, where $\by_{j}^{(i)}$ spans the same
temporal interval as $\bx_{j}^{(i)}$.
Given these group
trajectories, the individual trajectory segments are mutually
independent, so that
\begin{align}
\label{eq:individual-likelihood}
p(\by_j \given &\bX_{j}, \bC_{(j)}) =
    \prod_{i=1}^{I_j} p(\by_j^{(i)} \given \bx_{j}^{(i)}, a_{(j)}^{(i)}),
\end{align}
where $a_{(j)}^{(i)}$ is the label of the $i$th activity in
sequence $\bC_{(j)}$.

We also use a Gaussian process for individual trajectories, but fix the mean
to the (given) group trajectory, i.e.
$\by_j^{(i)} \sim \mathcal{GP}(\bx_j^{(i)}, \kappa)$, which implies that
$\by_j^{(i)} \sim \Norm(\bx_j^{(i)}, \mathbf{K}_j^{(i)})$,
where the $ij$th entry in $\mathbf{K}_j^{(i)}$ is the
covariance function $\kappa$
evaluated at the $i$th and $j$th frames of $\by_j^{(i)}$, using the
scale corresponding to the activity label associated to $\bx_j^{(i)}$.
See Figure \ref{fig:act-traj} for an example graphical model of
this distribution.

\subsection{Specific Activities}
\label{sec:specific-activities}

In this work, we limit ourselves to six specific activities, three
intentional ({\sc free-for-all}, {\sc meet}, and {\sc move-to})
and three physical ({\sc stand}, {\sc walk}, and {\sc run}).
{\sc FFA} has a single role which allows all activities to take place.
A {\sc meet} activity assigns
non-zero probability to two roles, {\sf APPROACHER} and
{\sf WAITER}; an {\sf APPROACHER} performs
a {\sc meet}s (recursively) and {\sc move-to}s, and
a {\sf WAITER} only performs {\sc stand}s.
{\sc Move-to} only produces one role, {\sf MOVER}, which switches
uniformly at random between the three physical activities.
Finally, the physical activities have scale parameters such that
$\sigma_{\textrm{STAND}} < \sigma_{\textrm{WALK}} < \sigma_{\textrm{RUN}}$.


\begin{figure*}[tb!]
\centering
\subfigure[Birth/death]{
    \includegraphics[width=0.22\linewidth]{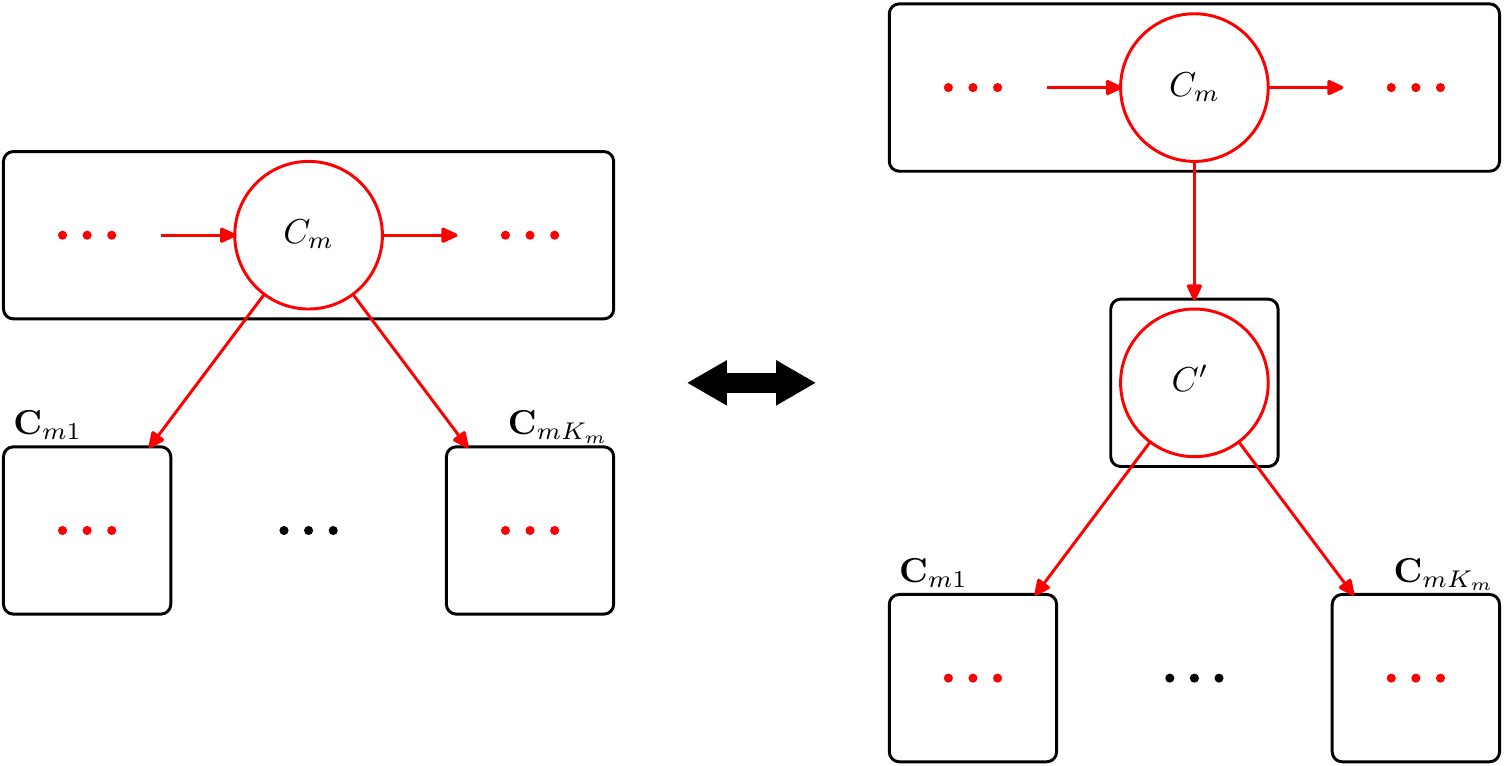}
    \label{fig:birth-move}
}
\hspace{5mm}
\subfigure[Merge/split]{
    \includegraphics[width=0.22\linewidth]{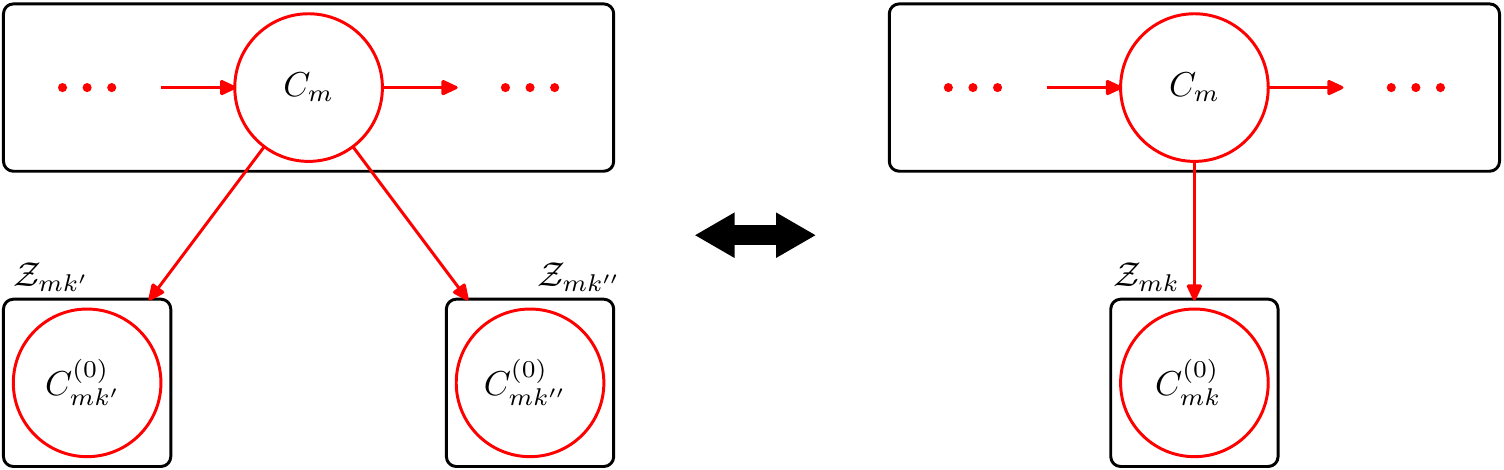}
    \label{fig:merge-move}
}
\hspace{5mm}
\subfigure[Sequence/unsequence]{
    \includegraphics[width=0.44\linewidth]{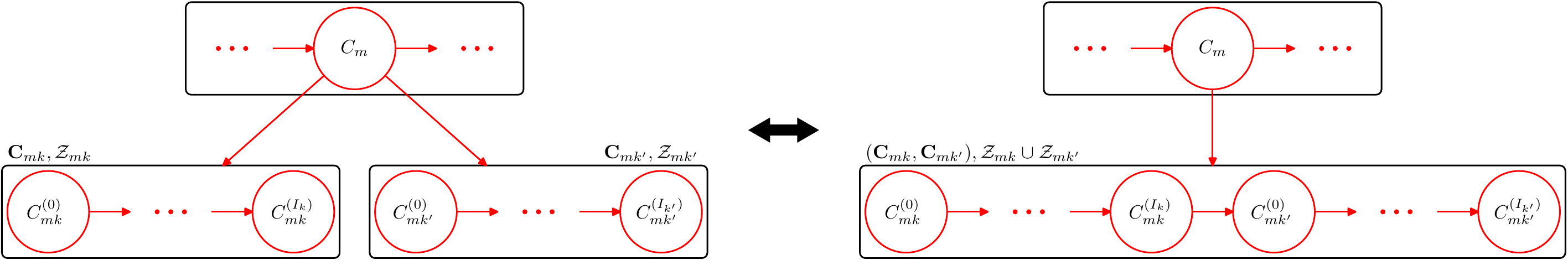}
    \label{fig:seq-move}
}
\caption{An illustration of our sampling moves. \subref{fig:birth-move}~
    The birth move (left to right) inserts an intentional activity
    $C'$ between sequences $\bC_{m1}, \dots, \bC_{m K_m}$ and their parent
    $C_m$. In the death move, the opposite operation is performed.
    \subref{fig:merge-move}~
    In the merge move (left to right), children $C_{mk'}^{(0)}$ and
    $C_{mk''}^{(0)}$ of intentional activity $C_m$ (and
    members of single-activity sequences), with
    $a_{mk'}^{(0)} = a_{mk''}^{(0)}$, are replaced by
    a single activity node $C_{mk}^{(0)}$, with activity
    $a_{mk'}^{(0)}$, and participant set
    $\cZ_{mk'} \cup \cZ_{mk''}$.
    In the reverse move, $C_{mk}^{(0)}$
    is split into two nodes, both with activity label
    $a_{mk}^{(0)}$, with $\cZ_{m}$ randomly partitioned.
    \subref{fig:seq-move}~ In this example, a sequence move (left to
    right)
    takes two child sequences of intentional activity node $C_m$,
    $\bC_{mk}$ and $\bC_{mk'}$, and forms a single sequence
    $(\bC_{mk}, \bC_{mk'})$ with participant set
    $\cZ_{mk} \cup \cZ_{mk'}$. In the reverse move, we randomly choose
    a split point in the activity sequence on the right, and create
    two separate sequences under the same parent $C_m$.}
    \label{fig:moves}
\end{figure*}

\section{Inference}
\label{sec:inference}

Given a set of $J$ individual trajectories $\bY = (\by_1, \dots, \by_J)$, 
such as those depicted in Figure \ref{fig:ex-trajs-3},
we wish to find an activity tree $\Lambda$, 
such as that depicted in Figure \ref{fig:ex-tree},
that best describes them. Specifically,
we wish to maximize the posterior probability of an activity tree given the
observed data
\begin{equation}
    p(\Lambda \given \bY) \propto p(\Lambda) p(\bY \given \Lambda),
        \label{eq:posterior}
\end{equation}
where the prior is given
by \eqref{eq:tree-prior},
and the likelihood is
\begin{equation}
    p(\bY \given \Lambda) = \int p(\bX \given \Lambda) \prod_{j=1}^J
        p(\by_j \given \bX_j, \bC_j) \, \textrm{d} \bX.
        \label{eq:likelihood-full}
\end{equation}
The integrands are given by \eqref{eq:physical-traj-prior} and
\eqref{eq:individual-likelihood}. In general, the integral in
\eqref{eq:likelihood-full} cannot be computed analytically. However, since every
factor in \eqref{eq:likelihood-full} is a normal pdf, $p(\bY \given \Lambda)$
is also normal, which makes the evaluation of \eqref{eq:posterior}
straightforward.

We cannot find $\Lambda^* = \argmax_{\Lambda} p(\Lambda \given \bY)$
analytically. Instead, we draw $S$ samples from the posterior
\eqref{eq:posterior} using the Metropolis-Hastings (MH) algorithm,
and keep the sample with the highest posterior probability.  At the
$i$th iteration, we draw $\Lambda'$ from a proposal distribution $q(\cdot
\given \Lambda^{(i-1)})$, where $\Lambda^{(i-1)}$ is the current sample, and
accept the sample with the standard MH acceptance probability
\cite{Neal-93-MCMC}.

The ability of MH to efficiently explore the
space depends largely on the choice of the proposal distribution $q$.
Although there has been work on MCMC sampling on tree models
\cite{bayesian-treed-chipman,pratola-mh-tree}
there is no general approach which can be applied to any model.
Consequently, we employ a proposal distribution which is specific 
to our model.

\subsection{Proposal distribution}
\label{sec:proposal-distro}

Our proposal mechanism is composed of
sampling moves which perform edits to the 
current hypothesized activity tree to produce a new tree sample.
When drawing a sample from $q$, we choose a move uniformly at random
to apply. When applying a move, we must make sure that the
resulting tree is valid (e.g., start and end times must be consistent;
or activity sequences must be possible given the role),
which requires 
book-keeping that is beyond the scope of this document.


We have also developed a set of bottom-up activity detectors to help
explore the space efficiently. These detectors provide rough estimates of
groupings
of individuals at each frame, and activities being performed
by each group (see Section \ref{sec:detectors}, ``Detectors''). We use
these detectors in two ways. First, we initialize the sampler
to a state obtained by transforming the output of the detectors to
an activity tree $\Lambda^{(0)}$. Additionally, we bias our proposal
distribution toward groups and activities found by the detectors.
For example, when proposing a \emph{merge} move, we might choose
participants which are predicted to be in a group by some activity
detector.

\paragraph{Sampling moves}
During inference, we employ the following moves (see Figure \ref{fig:moves}
for an illustration).
(a) {\bf Birth/death}:
a birth move inserts an intentional activity node
between an intentional
activity and some of its children.
We randomly choose a set of sibling activity sequences
$\bC_{m1}, \dots, \bC_{m K_m}$ whose parent is intentional activity $C_m$,
and insert a new intentional activity node $C'$ (whose label is also chosen
at random),
such that $C'$ becomes the
parent of $\bC_{m1}, \dots, \bC_{m K_m}$ and a child of $C_m$.
In a death move, we randomly choose a intentional node $C'$, remove it from
the tree, and connect its children nodes to its parent.
(b) {\bf Merge/split}:
In a merge move, we take two sibling activity nodes with the same label and
combine them into a single activity. If $C'$ and $C''$ are two activities with
label $a$ and groups $\cZ'$ and $\cZ''$, we create a new node $C$ with
label $a$ and participants $\cZ' \cup \cZ''$. The split move performs
the opposite operation, taking $a$-labeled node $C$ and splitting it into
two nodes $C'$ and $C''$, both with activity $a$, assigning
participants in $\cZ$ to either $\cZ'$ or $\cZ''$ uniformly.
(c) {\bf Sequence/unsequence}:
Let $\bC_1$ and $\bC_2$ be two temporally non-overlapping sibling activity
sequences. A sequence move concatenates
$\bC_1$ and $\bC_2$ into a new sequence $\bC$. An unsequence move
randomly selecting a split point at which to
separate a sequence.
(d) {\bf Relabel move}:
The relabel move randomly changes $C$'s label. Note that the
label must be valid, e.g.,
we cannot assign a physical activity label to an intentional activity node.

\subsubsection{Detectors}
\label{sec:detectors}

The bottom-up detectors provide an estimate of how individuals in the
video are grouped across time, as well as the physical activity they
are performing.

At each frame, we cluster individuals into groups using their trajectories
on the ground plane.
We apply the density-based spatial clustering of applications with noise
(DBSCAN) \cite{Ester96adensity-based} algorithm independently on each frame,
where our feature is composed of the position and velocity of an individual
at that frame, both of which are obtained from the smoothed trajectories
(which are assumed to be given).
Importantly, we keep track of individual identities
over time by recording the actors involved in each group in the previous frame
and
assigning the cluster found in the following frame where the majority of
individuals in that new set are still involved.
Given the groups as computed above, we want to identify the physical
activities of their individuals (e.g., {\sc walk, run}). 
For this we use a hidden Markov model, where the observation function is a naive Bayes model with each individual's speed modeled by a Gamma distribution, and a transition function that prefers staying within the current activity.

\section{Experiments and Results}
\label{sec:experiments}

We evaluate the model in two ways.  First, we demonstrate the model's
expressive power in inferring two different types of complexly structured
scenarios from synthetic data. In the first, groups of individuals engage in activities and
disband, forming different groups over time. The second demonstrates a
recursively structured activity, in which one meeting is a component of a
higher-level meeting. We then evaluate the model on real data;
specifically on two publicly available
group activity datasets, VIRAT \cite{42_oh2011large} and the
UCLA aerial event dataset \cite{shu2015_CVPR}.


\begin{figure}[tb]
\centering
\subfigure[VIRAT frames]{
    \includegraphics[width=0.48\linewidth]{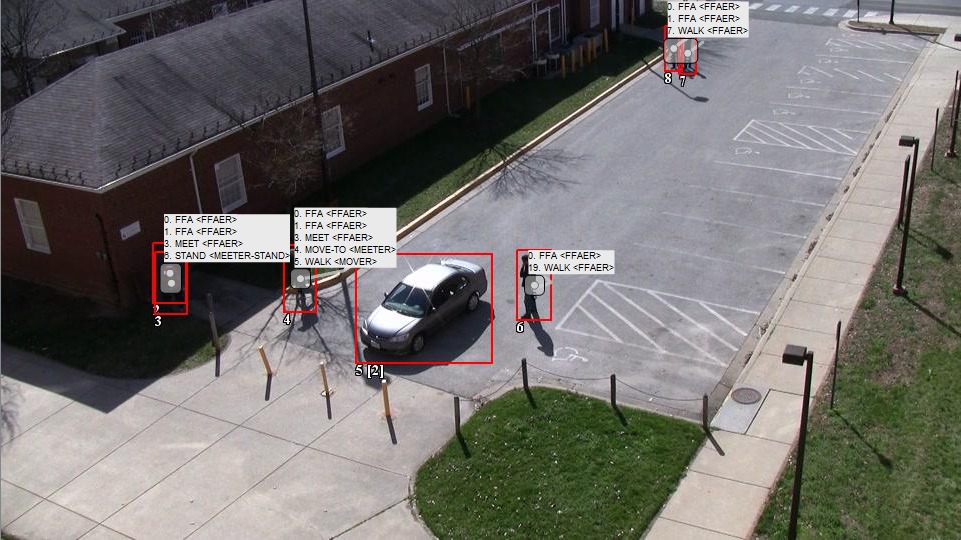}
    \includegraphics[width=0.48\linewidth]{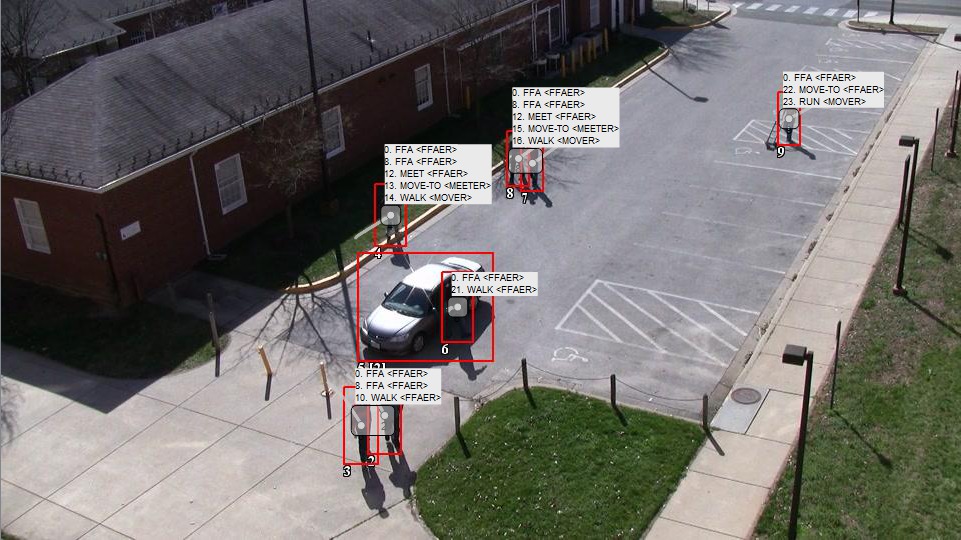}
    \label{fig:virat-frms}
}
\subfigure[Inferred activity tree]{
    \includegraphics[width=\linewidth]{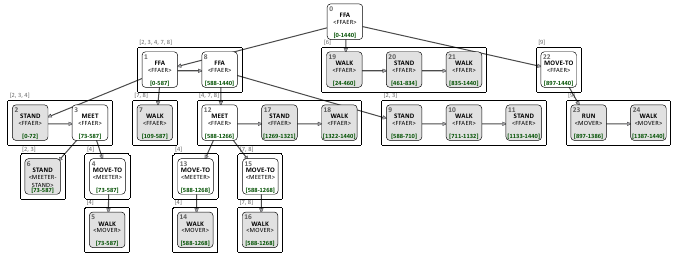}
    \label{fig:virat-tree}
}
\caption{Qualitative results the VIRAT dataset. \subref{fig:virat-frms}~
    shows two frames of the VIRAT2 video. 
    Red boxes represent tracked individuals,
    gray rectangles indicate groups of
    individuals, and white boxes contain information such as
    track id, etc. In
    \subref{fig:virat-tree}~ we show the inferred activity tree,
    illustrated in a similar way as in Figure \ref{fig:ex-tree}.}
\label{fig:virate-qual}
\end{figure}

\subsection{Evaluation}
\label{sec:evaluation}

Performance is measured in terms of how well activities are labeled
in the scene, and how well individuals are grouped, irrespective of
activity label. In the following, $\Lambda_{gt}$ and $\Lambda$ are
the ground truth and inferred activity trees, respectively.

\subsubsection{Activity Labeling} 

For each activity $a$ and video frame $f$, we compare between
$\Lambda_{gt}$ and $\Lambda$ the set of individuals performing $a$ at $f$.
We first define performance counts in terms of an individual at a frame, then
compute overall counts and associated performance measures. 

For an individual $j$ at frame $i$, let $\lambda_{ji}$
be the set of individuals which have the same label as $j$ in
$\Lambda$. Define $\lambda_{ji}'$ similarly for
$\Lambda_{gt}$. The set of false positives for $j$ is
$\lambda_{ji}\setminus\lambda_{ji}'$, the set of false
negatives is $\lambda_{ji}'\setminus\lambda{ji}$,
the true positives are
$\lambda_{ji}'\cap\lambda_{ji}$, and the true
negatives are $(\mathcal{Z}\setminus\lambda_{ji}) \cap
(\mathcal{Z}\setminus\lambda_{ji}')$, where $\mathcal{Z}$ is
the set of all individuals.

\subsubsection{Grouping}

We follow a similar approach when evaluating grouping performance.
For person $j$, we compare the groups to
which $j$ belongs in $\Lambda_{gt}$ and $\Lambda$ at each frame
$t$. Since the two trees could have different depths and topologies,
it is not necessarily clear which groups should be compared with which;
however, every individual is part of exactly one physical activity
group at each frame, as well as one highest level node. Consequently,
we only compare groups at these two levels of the tree, without penalty
for difference in activities within a level. Thus once we determine
the group associated with person $j$ at frame $t$ at the physical
(resp. highest) activity level of each tree, we compute
the score as before. 


\makeatletter
\providecommand{\tabularnewline}{\\}
\makeatother

\subsection{Results}

The performance of our algorithm is summarized in Tables \ref{tab:act-label}
and \ref{tab:group-performance}. Table \ref{tab:act-label} shows the
activity labeling precision and recall on the two synthetic scenes
(SYNTH1 and SYNTH2),
a video sequence obtained from the VIRAT dataset, and four different
video sequences from the UCLA aerial event dataset (UAED).
In Table \ref{tab:group-performance} we see the performance as measured
by our grouping evaluation metric described above.

\begin{table}[tb]
\setlength\tabcolsep{4.5pt}
\scriptsize
\centering
\begin{tabular}{|c|c|c|c|c|c|}
\hline
{\textbf{SYNTH1}} & STAND & WALK & MOVE-TO & MEET & FFA \tabularnewline
\hline 
Precision & 0.59 & 0.84 & 0.63 & 0.71 & 1.00 \tabularnewline
\hline 
Recall & 1.00 & 0.94 & 0.71 & 0.75 & 1.00 \tabularnewline
\hline 
F1 & 0.74 & 0.88 & 0.67 & 0.73 & 1.00 \tabularnewline
\hline 
\hline
{\textbf{SYNTH2}} & STAND & WALK & MOVE-TO & MEET &FFA \tabularnewline
\hline 
Precision & 0.87 & 1.00 & 0.94 & 1.00 & $\times$ \tabularnewline
\hline 
Recall & 0.50 & 0.71 & 0.73 & 0.56 & $\times$ \tabularnewline
\hline 
F1 & 0.63 & 0.83 & 0.82 & 0.72 & $\times$ \tabularnewline
\hline 
\hline
{\textbf{VIRAT}} & STAND & WALK & MOVE-TO & MEET & FFA \tabularnewline
\hline 
Precision & 0.85 & 0.81 & 0.73 & 0.90 & 1.0 \tabularnewline
\hline 
Recall & 0.51 & 0.88 & 0.89 & 0.80 & 1.0 \tabularnewline
\hline 
F1 & 0.64 & 0.85 & 0.81 & 0.85 & 1.0 \tabularnewline
\hline 
\hline
{\textbf{UAED}} & STAND & WALK & MOVE-TO & MEET & FFA \tabularnewline
\hline 
Precision & 0.96 & 0.89 & 0.71 & 0.75 & 0.97 \tabularnewline
\hline 
Recall & 0.82 & 0.99 & 0.67 & 0.64 & 0.73 \tabularnewline
\hline 
F1 & 0.86 & 0.94 & 0.62 & 0.62 & 0.77 \tabularnewline
\hline 
\end{tabular}
\caption{Activity labeling results for synthetic videos
    SYNTH1 and
    SYNTH2, and the
    VIRAT and
    UCLA aerial event datasets. Each table shows
    precision, recall, and F1 for each activity.
    See Section \ref{sec:evaluation} for details.}
\label{tab:act-label}
\end{table}

\subsubsection{Synthetic data}

The synthetic dataset comprises two videos, where a video is
a set of trajectories on the ground plane. In the first,
SYNTH1, five actors participate in a series of meetings,
where participants repeatedly change group memberships across 20 frames.
The second (SYNTH2) features five actors meeting, with
four of them participating in a side meeting before joining the
global meeting. As Table
\ref{tab:act-label}
shows, our model performs very well
on high-level activities, such as {\sc meet}, even when presented with
nested structure.

\subsubsection{VIRAT}

We also evaluate on real data, specifically frames
2520 to 3960 of video 2 of the VIRAT dataset.
This video features seven people participating in two meetings, where groups
exchange members several
times. Fig. \ref{fig:virate-qual} shows two frames of the video, along
with the inferred activity tree. Our model correctly recognizes
the two meetings, as well as all of the groups at the highest level
of description. 
The activity labeling results (Table
\ref{tab:act-label})
show perfect {\sc FFA} performance, and the grouping results
(Table \ref{tab:group-performance}) show a perfect
highest-level intentional activity score. As before, there is divergence
from ground truth at the physical activity level, but this does not
affect the grouping score.

\begin{table}[tb]
\setlength\tabcolsep{4.5pt}
\scriptsize
\centering
\begin{tabular}{|c|c|c|c|c|c|c|c|c|}
    \cline{2-9} 
    \multicolumn{1}{c|}{} & \multicolumn{2}{c|}{\textbf{SYNTH1}} & \multicolumn{2}{c|}{\textbf{SYNTH2}} & \multicolumn{2}{c|}{\textbf{VIRAT}} & \multicolumn{2}{c|}{\textbf{UAED}} \tabularnewline
    \multicolumn{1}{c|}{} & PHYS  & INT  & PHYS  & INT & PHYS & INT & PHYS & INT \tabularnewline
    \hline 
    Precision & 0.86 & 1.0 & 1.00 & 1.0 & 0.98 & 1.0 & 0.95 & 0.93 \tabularnewline
    \hline 
    Recall & 1.00 & 1.0 & 0.69 & 1.0 & 0.85 & 1.0 & 0.99 & 0.89 \tabularnewline
    \hline 
    F1 & 0.92 & 1.0 & 0.82 & 1.0 & 0.91 & 1.0 & 0.97 & 0.89 \tabularnewline
    \hline 
\end{tabular}
\caption{Grouping precision, recall, and F1 scores for the
    synthetic videos and the two datasets
    VIRAT and UCLA aerial event dataset (UAED).
    See Section \ref{sec:evaluation} for details.}
\label{tab:group-performance}
\end{table}

\subsubsection{UCLA aerial event dataset}

We extracted four video sequences from the UCLA aerial event dataset (UAED).
More specifically, we searched for subsequences of videos which featured
properties like activity nesting, groups interchanging members, etc. The
result is four video sequences of roughly $2000$ frames each.
As we can see in Table
\ref{tab:act-label},
which shows the overall precision and recall scores, for all four
videos, our algorithm performs reasonably well across all
activities. Note the relatively low recall score of the {\sc meet}
activity, which is due in large part to one very long missed
{\sc meet} in the third video sequence.
Similarly, Table \ref{tab:group-performance} shows that our algorithm
performs well at finding groups of individuals at both the physical
and intentional activity levels.


\section{Discussion}

We have presented a probabilistic generative model of complex multi-agent
activities over arbitrary time scales.
The activities specify component roles between groups of actors and accommodate
unboundedly deep recursive, hierarchical structure.  The model accommodates
arbitrary groups participating in activity roles, describing both between-group
and between-individual interactions.  Physical and intentional (higher-level
description) activities explain hierarchical correlations among individual
trajectories.  To our knowledge, no existing model of track-based activity
recognition provides this expressiveness in a joint model.  

The modeling framework is naturally extensible.  We are currently undertaking
several extensions, including 
(1) developing additional activities,
including following, exchanging items, and interacting with vehicles and
building entrances, 
(2) adding prior knowledge about the spatial layout of the scene
that naturally constrains what activities are possible, such as roads,
sidewalks, impassible buildings, and other spatial features that
influence behavior in order to improve both accuracy and
speed by reducing the search space,
(3) using our model as a prior for a 3D Bayesian tracker
\cite{brau2013},
and (4) connecting natural language to activity
descriptions 
as our model accommodates activity descriptions across
multiple events, tracking individual participation throughout, providing
opportunities for building natural language narratives about activities at
different levels of granularity.


\section*{Acknowledgements}
This research was supported by grants under the DARPA Mind's Eye program
W911NF-10-C-0081 (subcontract to iRobot, 92003) and the DARPA SSIM program
W911NF-10-2-0064.  We give special thanks to Paul R. Cohen and Christopher
Geyer for helpful discussions and advice.

\small{
\bibliography{uab3-bibliography}
\bibliographystyle{aaai}
}

\end{document}